\documentclass{article} 
\usepackage{iclr2020_conference,times}

\usepackage{amsmath,amsfonts,bm}

\def\eqref#1{equation~\ref{#1}}

\def\1{\bm{1}}

\DeclareMathAlphabet{\mathsfit}{\encodingdefault}{\sfdefault}{m}{sl}
\SetMathAlphabet{\mathsfit}{bold}{\encodingdefault}{\sfdefault}{bx}{n}

\usepackage{hyperref}
\usepackage{url}

\usepackage[utf8]{inputenc} 
\usepackage[T1]{fontenc}    
\usepackage{booktabs}       
\usepackage{amsfonts}       
\usepackage{nicefrac}       
\usepackage{microtype}      
\usepackage{latexsym}
\usepackage{url}
\usepackage{color}

\usepackage{graphicx}
\usepackage{amsmath}

\newcommand{\trm}[1]{\textit{{#1}}}

\newcommand{\myvspace}[1]{\vspace{#1}}

\newtheorem{claim}{Claim}

\newcommand{\aq}{q}
\newcommand{\aA}{A}
\newcommand{\aex}{(\aq,\aA)}
\newcommand{\arel}{r}
\newcommand{\aent}{x}
\newcommand{\srel}{R}
\newcommand{\sent}{X}

\newcommand{\followop}{relation-set following}
\newcommand{\follow}[2]{\textit{follow}({#1},{#2})}
\newcommand{\ty}{\textit{type}}

\newcommand{\aty}{\tau}
\newcommand{\aR}{\mathbb{R}}
\newcommand{\vek}[1]{\textbf{#1}}
\newcommand{\aind}{index}
\newcommand{\subj}{{\it subj}}
\newcommand{\obj}{{\it obj}}
\newcommand{\rel}{{\it rel}}

\newcommand{\setw}[2]{\omega|\![{#1} \in {#2}]\!|}

\newcommand{\supp}{\textit{support}}

\title{Scalable Neural Methods for \\Reasoning With a Symbolic Knowledge
  Base}

\author{%
  William W. Cohen \& Haitian Sun \& R. Alex Hofer \& Matthew Siegler\\
  Google, Inc
  \\ \texttt{\{wcohen,haitiansun,rofer,msiegler\}@google.com} \\
}

\newif\ifstilldraft \stilldraftfalse
\iclrfinalcopy 

\begin{document}

\maketitle

\begin{abstract}
  We describe a novel way of representing a symbolic knowledge base
  (KB) called a \textit{sparse-matrix reified KB}.  This representation enables
  neural KB inference modules that are fully differentiable, faithful to the
  original semantics of the KB, expressive enough to model multi-hop
  inferences, and scalable enough to use with realistically large
  KBs. The sparse-matrix reified KB can be distributed across multiple
  GPUs, can scale to tens of millions of entities and facts, and is
  orders of magnitude faster than naive sparse-matrix
  implementations.  The reified KB
  enables very simple end-to-end architectures to obtain competitive
  performance on several benchmarks representing two families of
  tasks: KB completion, and learning semantic parsers from
  denotations.
\end{abstract}

\myvspace{-0.1in}

\section{Introduction}

\myvspace{-0.1in}

There has been much prior work on using neural networks to
\emph{generalize} the contents of a KB 
\citep{xiong2017deeppath,bordes2013translating,dettmers2018convolutional},
typically by constructing low-dimensional embeddings of the entities
and relations in the KB, which are then used to score potential
triples as plausible or implausible elements of the KB.  We consider
here the related but different problem of \emph{incorporating a
  symbolic KB into a neural system}, so as to inject knowledge from
an existing KB directly into a neural model.  More precisely, we consider the
problem of designing neural KB inference modules that are (1) fully differentiable, so that any loss based on their outputs
can be backpropagated to their inputs; (2) accurate, in that they are
faithful to the original semantics of the KB; (3) expressive, so they can perform non-trivial inferences; and (4) scalable, so that realistically large KBs can be incorporated into a neural model.

To motivate the goal of incorporating a symbolic KB into a neural
network, consider the task of learning neural semantic parsers from
denotations.  Many questions---e.g., \textit{what's the most recent
  movie that Quentin Tarantino directed?}~or \textit{which nearby
  restaurants have vegetarian entrees and take reservations?}---are
best answered by \trm{knowledge-based question-answering} (KBQA)
methods, where an answer is found by accessing a KB.  Within KBQA, a
common approach is \trm{neural semantic parsing}---i.e., using neural
methods to translate a natural-language question into a structured
query against the KB 
\citep{zhong2017seq2sql,finegan2018improving,shaw2019generating},
which is subsequently executed with a symbolic KB query engine.  While
this approach can be effective, it requires training data pairing
natural-language questions with structured queries, which is difficult
to obtain.  Hence researchers have also considered \trm{learning
  semantic parsers from denotations} 
\citep{berant2013semantic,yih2015semantic}, where training data
consists of pairs $\aex$, where $\aq$ is a natural-language question
and $\aA$ is the desired answer. Typically $\aA$ is a set of KB
entities---e.g., if $q$ is the first sample question above, $\aA$
would be\footnote{At the time of this writing.} the singleton set
containing \textit{Once Upon a Time in Hollywood}.

Learning semantic parsers from denotations is difficult because the
end-to-end process to be learned includes a non-differentiable
operation---i.e., reasoning with the symbolic KB that contains the
answers. To circumvent
this difficulty,  prior systems have used three different approaches.  Some have used heuristic search to infer
structured queries from denotations
\citep{pasupat2016inferring,dasigi-etal-2019-iterative}: this works in some cases but often an answer could be associated with many possible structured queries, introducing noise. Others have
supplemented gradient approaches with reinforcement learning 
(e.g., \citep{misra2018policy}).  Some systems have also ``neuralized'' KB reasoning, but to date only over
\emph{small} KBs: this approach is natural when answers are naturally constrained to 
depend on a small set of facts (e.g., a single table
\citep{zhong2017seq2sql,DBLP:journals/corr/abs-1808-09942}), but more generally requires coupling a learner with some (non-differentiable) mechanism to retrieve an appropriate small question-dependent subset of the KB as in  \citep{graftnet,sun2019pullnet}.

In this paper, we introduce a novel scheme for incorporating reasoning
on a large question-independent KB into a neural network, by
representing a symbolic KB with an encoding called a
\trm{sparse-matrix reified KB}.  A sparse-matrix reified KB is very
compact, can be distributed across multiple GPUs if necessary, and is
well-suited to modern GPU architecture. For KBs with many relations, a
reified KB can be up to four orders of magnitude faster than
alternative implementations (even alternatives based on sparse-matrix
representations), and in our experiments we demonstrate scalability to
a KB with over 13 million entities and nearly 44 million facts.  This
new architectural component leads to radically simpler architectures
for neural semantic parsing from denotations---architectures based on
a single end-to-end differentiable process, rather than cascades of
retrieval and neural processes.

We show that
very simple instantiations of these architectures are still highly competitive with the
state of the art for several benchmark tasks.  \emph{To our knowledge these models are the first fully end-to-end neural parsers from denotations that have been applied to these benchmark tasks.} We also demonstrate
that these architectures scale to long chains of reasoning on
synthetic tasks, and demonstrate similarly simple architectures for a
second task, KB completion.

\myvspace{-0.1in}

\section{Neural reasoning with a symbolic KB}

\myvspace{-0.1in}

\subsection{Background}

\myvspace{-0.04in}

\begin{table}[t]
\begin{small}
    \centering
    \begin{tabular}{llll}
      $\aent$: an entity    &   $\sent$: weighted set of entities & $\vek{\aent}$: vector encoding $\sent$  & $N_E$: \# entities in KB \\
      $\arel$: an relation  &   $\srel$: weighted set of relations & $\vek{\arel}$: vector encoding $\srel$ & $N_R$: \# relations in KB \\
      $\vek{M}_\arel$: matrix for $\arel$  
                            & $\vek{M}_\srel$: weighted sum of $\vek{M}_\arel$'s, see Eq~\ref{eq:MR} 
                            & $\follow{\vek{\aent}}{\vek{\arel}}$: see Eq~\ref{eq:followdef}
                            & $N_T$: \# triples in KB \\
      \multicolumn{4}{l}{$\vek{M}_\subj, \vek{M}_\obj, \vek{M}_\rel$: the reified KB, encoded as matrices mapping triple id $\ell$ to subject, object, and relation ids}
    \end{tabular}
\end{small}
\caption{Summary of notation used in the paper.  (This excludes notation used in defining models for the KB completion and QA tasks of Section~\ref{sec:expt}.)}
\end{table}

\textbf{KBs, entities, and relations.}  A KB consists of
\trm{entities} and \trm{relations}.  We use $\aent$ to denote an
entity and $\arel$ to denote a relation.  Each entity has an integer
index between $1$ and $N_E$, where $N_E$ is the number of entities in
the KB, and we write $\aent_{i}$ for the entity that has index $i$.  A
relation is a set of entity pairs, and represents a relationship
between entities: for instance, if $\aent_i$ represents ``Quentin
Tarantino'' and $\aent_j$ represents ``Pulp Fiction'' then
$(\aent_i,\aent_j)$ would be an member of the relation
\textit{director\_of}.  A relation $\arel$ can thus be represented as a subset of \(
\{1,\ldots,N_E\} \times \{1,\ldots,N_E\} \).  Finally a KB consists a
set of relations and a set of entities.

\textbf{Weighted sets as ``$k$-hot'' vectors.} Our differentiable
operations are based on \trm{weighted sets}, where each element
$\aent$ of weighted set $\sent$ is associated with a non-negative real
number.  It is convenient to define this weight to be zero for all
$\aent\not\in\sent$, while for $\aent\in\sent$, a weight less than 1 is
a confidence that the set contains $\aent$, and weights more than 1
make $\sent$ a multiset.  If all elements of $\sent$ have weight 1, we
say $\sent$ is a \trm{hard set}. A weighted set $\sent$ can be encoded
as an \trm{entity-set vector} $\vek{\aent} \in \aR^{N_E}$, where the
$i$-th component of $\vek{\aent}$ is the weight of $x_i$ in
$\sent$. If $\sent$ is a hard entity set, then this will be a
``$k$-hot'' vector, for $k=|\sent|$.  The set of indices of
$\vek{\aent}$ with non-zero values is called the \trm{support of
  $\vek{\aent}$}.

\textbf{Sets of relations, and relations as matrices} Often we would
like to reason about sets of relations\footnote{This is usually called
  \trm{second-order} reasoning.}, so we also assume every relation
$\arel$ in a KB is associated with an entity and hence an integer
index.  We write $\arel_k$ for the relation with index $k$, and we
assume that relation entities are listed first in the index of
entities, so the index $k$ for $r_k$ is between 1 and $N_R$, where
$N_R$ is the number of relations in the KB.  We use $\srel$ for a set
of relations, e.g., $\srel = \{ \textit{writer\_of},
\textit{director\_of} \}$ might be such a set, and use $\vek{\arel}$
for a vector encoding of a set.  A relation $\arel$ can be encoded as
a \trm{relation matrix} $\vek{M}_\arel \in \aR^{N_{E} \times N_{E}}$,
where the value for $\vek{M}_\arel[i,j]$ is (in general) the weight of
the assertion $\arel(\aent_i,\aent_j)$ in the KB.  In the experiments
of this paper, all KB relations are hard sets, so
$\vek{M}_\arel[i,j]\in\{0,1\}$.

\textbf{Sparse vs.~dense matrices for relations}.  Scalably
representing a large KB requires careful consideration of the
implementation.  One important issue is that for all but the smallest
KBs, a relation matrix must be implemented using a \trm{sparse matrix}
data structure, as explicitly storing all $N_E^2$ values is
impractical.  For instance, consider a KB containing 10,000 movie
entities and 100,000 person entities.  A relationship like
\textit{writer\_of} would have only a few tens of thousands of facts (since most movies have only one or two writers), but a dense matrix
would have 1 billion values.

We thus model relations as \trm{sparse matrices}.  Let $N_\arel$ be
the number of entity pairs in the relation $\arel$: common sparse matrix data structures require space $O(N_\arel)$. One common sparse
matrix data structure is a \trm{sparse coordinate pair (COO)}
encoding: with a COO encoding, each KB fact requires storing only
two integers and one float.

Our implementations are based on Tensorflow
\citep{abadi2016tensorflow}, which offers limited support for sparse
matrices.  In particular, driven by the limitations of GPU
architecture, Tensorflow only supports matrix multiplication between a
sparse matrix COO and a dense matrix, but not between two sparse matrices, or between sparse higher-rank tensors and dense tensors.

\textbf{Entity types.}  It is often possible to easily group entities
into disjoint sets by some notion of ``type'': for example, in a movie
domain, all entities might be either of the type ``movie'',
``person'', or ``movie studio''.  It is straightforward to extend the
formalism above to typed sets of entities, and doing this can lead to
some useful optimizations. 
We use these optimizations below where appropriate: in particular, relation-set vectors $\vek{\arel}$ are of dimension
$N_R$, not $N_E$, in the sections below.  The full formal extension to typed entities and relations is given in
Appendix~\ref{sec:types}.

\myvspace{-0.04in}

\subsection{Reasoning in a KB}

\myvspace{-0.03in}

\textbf{The \followop{} operation.}  Note that relations can also be viewed as labeled
edges in a \trm{knowledge graph}, the vertices of which are entities.
Adopting this view, we define the \trm{$\arel$-neighbors of an entity
  $\aent_i$} to be the set of entities $\aent_j$ that are connected to
$\aent_i$ by an edge labeled $\arel$, i.e., \(
\textit{$\arel$-neighbors($\aent$)} \equiv \{ \aent_j : (\aent_i,\aent_j)
\in \arel \} \).  Extending this to relation sets, we define
\[ \textit{$\srel$-neighbors($\sent$)} \equiv \{ \aent_j : \exists \arel\in\srel, \aent_i\in\sent \mbox{~so that~} (\aent_i,\aent_j) \in \arel \}
\]
Computing the $\srel$-neighbors of an entity is a single-step
reasoning operation: e.g., the answer to the question
$q=$``\textit{what movies were produced or directed by Quentin Tarantino}'' is
precisely the set $\srel$-neighbors($\sent$) for
$\srel=\{\textit{producer\_of, writer\_of}\}$ and
$\sent=\{\textit{Quentin\_Tarantino}\}$. ``Multi-hop'' reasoning
operations require nested $\srel$-neighborhoods, e.g.  if
$\srel'=\{\textit{actor\_of}\}$ then
$\srel'$-{neighbors}($\srel$-{neighbors}$(\sent))$ is the set of
actors in movies produced or directed by Quentin Tarantino.

We would like to approximate the $\srel$-neighbors computation with
differentiable operations that can be performed on the vectors
encoding the sets $\sent$ and $\srel$.  Let $\vek{\aent}$ encode a
weighted set of entities $\sent$, and let $\vek{\arel}$ encode a
weighted set of relations.  We first define $\vek{M}_\srel$ to be a
weighted mixture of the relation matrices for all relations in $\srel$
i.e.,
\begin{equation} \label{eq:MR}
\vek{M}_\srel \equiv (\sum_{k=1}^{N_R}
\vek{\arel}[k] \cdot \vek{M}_{r_k}) 
\end{equation}
We then define the \trm{\followop{} operation for
  $\vek{\aent}$ and $\vek{\arel}$} as:
\begin{equation}
 \label{eq:followdef}
    \follow{\vek{\aent}}{\vek{\arel}}   \equiv
     \vek{\aent} \vek{M}_\srel 
     = \vek{\aent} ( \sum_{k=1}^{N_{R}} \vek{\arel}[k] \cdot \vek{M}_{r_k} )
\end{equation}

\myvspace{-0.1in}

As we will show below, this differentiable numerical \followop{}
operation can be used as a neural component to perform certain types
of logical reasoning.  In particular, Eq~\ref{eq:followdef}
corresponds closely to the logical $\srel$-neighborhood operation, as
shown by the claim below.

\myvspace{-0.05in}

\begin{claim} The support of $\follow{\vek{\aent}}{\vek{\arel}}$ is
exactly the set of $\srel$-neighbors($\sent$).
\end{claim}

\myvspace{-0.05in}

A proof and the implications of this are discussed in
Appendix~\ref{sec:proof}.

\begin{table}[t]
    \centering
    \begin{tabular}{c|c|c|c|ccc}
    \hline
         Strategy & Definition    & Batch? & Space complexity & \multicolumn{3}{c}{\# Operations}\\
         & & & & sp-dense & dense  & sparse\\
         & & & & matmul   & + or $\odot$    & + \\

         \hline
         naive mixing     & Eq~\ref{eq:MR}-\ref{eq:followdef} & no  & $O(N_T + N_E + N_R)$     & 1 & 0 & $N_R$ \\
         late mixing      & Eq~\ref{eq:latemix} & yes & $O(N_T + b N_E + b N_R)$ & $N_R$ & $N_R$ & 0 \\
         reified KB       & Eq~\ref{eq:rjoin} & yes & $O(b N_T + b N_E)$       & 3     & 1     & 0 \\
         \hline
    \end{tabular}\\
    ~\\
    \caption{Complexity of implementations of \followop{}, where $N_T$ is the number of KB triples, $N_E$ the number of entities, $N_R$ the number of relations, and $b$ is batch size.}
    \label{tab:implementations}
    \vspace{-0.15in}
\end{table}

\subsection{Scalable \followop{} with a reified KB}

\textbf{Baseline implementations.}  Suppose the KB contains $N_R$
relations, $N_E$ entities, and $N_T$ triples.  Typically $N_R < N_E <
N_T \ll N_E^2$.  As noted above, we  implement each
$\vek{M}_\arel$ as a sparse COO matrix, so collectively these matrices
require space $O(N_T)$.  Each triple appears in only one relation, so
$\vek{M}_\srel$ in Eq~\ref{eq:MR} is also size $O(N_T)$.  Since
sparse-sparse matrix multiplication is not supported in Tensorflow we
implement $\vek{x}\vek{M}_\srel$ using dense-sparse multiplication,
so $\vek{x}$ must be a dense vector of size
$O(N_E)$, as is the output of \followop{}. Thus the space complexity of
$\follow{\vek{x}}{\vek{r}}$ is $O(N_T + N_E + N_R)$, if implemented as
suggested by Eq~\ref{eq:followdef}.  We call this the \trm{naive mixing} implementation,
and its complexity is summarized in Table~\ref{tab:implementations}.

Because Tensorflow does not support general sparse tensor
contractions, it is not always possible to extend sparse-matrix
computations to minibatches.  Thus we also consider a variant of naive
mixing called \trm{late mixing}, which mixes the \emph{output} of many
single-relation following steps, rather than mixing the KB itself:
\begin{equation}
 \label{eq:latemix}
    \follow{\vek{\aent}}{\vek{\arel}} = 
       \sum_{k=1}^{N_R} (\vek{\arel}[k] \cdot \vek{\aent} \vek{M}_{r_k})
\end{equation}
Unlike naive mixing, late mixing can be extended easily to a
minibatches (see Appendix~\ref{sec:mbatch}).  Let $b$ be the batch
size and $\vek{X}$ be a minibatch of $b$ examples $[\vek{x}_1; \dots; \vek{x}_b]$: then this approach leads to $N_R$ matrices $\vek{X} \vek{M}_k$,
each of size $O(b N_E$). However, they need not all be stored at once,
so the space complexity becomes $O(b N_E + b N_R + N_T)$.  An
additional cost of late mixing is that we must now sum up $N_R$ dense
matrices.

\textbf{A reified knowledge base.}
While semantic parses for natural questions often use small sets of relations (often singleton ones), in learning there is substantial uncertainty about what the members of these small sets should be.  Furthermore, realistic wide-coverage KBs have many relations---typically hundreds or thousands. This leads to a situation
where, at least during early phases of learning, it is necessary to evaluate the result of mixing very large sets of  relations. When many relations are mixed, late mixing becomes quite expensive (as experiments below show).

An alternative is to represent each KB assertion $r_k(x_i,x_j)$ as a
tuple $(i, j, k)$ where $i,j,k$ are the indices of $x_i,x_j$, and
$r_k$.  There are $N_T$ such triples, so for $\ell=1,\ldots,N_T$, let
\( (i_\ell, j_\ell, k_\ell) \) denote the $\ell$-th triple.
We define these sparse matrices:
$$
\vek{M}_\subj[\ell,m] \equiv \left\{ \!\!
   \begin{array}{l}
     \mbox{1~~if $m=i_\ell$} \\
     \mbox{0~~else} \\
   \end{array}
\right.~~~
\vek{M}_\obj[\ell,m] \equiv \left\{ \!\!
   \begin{array}{l}
     \mbox{1~~if $m=j_\ell$} \\
     \mbox{0~~else} \\
   \end{array}
\right.~~~
\vek{M}_\rel[\ell,m] \equiv \left\{ \!\!
   \begin{array}{l}
     \mbox{1~~if $m=k_\ell$} \\
     \mbox{0~~else} \\
   \end{array}
\right.
$$

Conceptually, $\vek{M}_\subj$ maps the index $\ell$ of the $\ell$-th
triple to its subject entity; $\vek{M}_\obj$ maps $\ell$ to the object
entity; and $\vek{M}_\rel$ maps $\ell$ to the relation. We can now implement the \followop{} as below, where
$\odot$ is Hadamard product:
\begin{equation}
 \label{eq:rjoin}
    \follow{\vek{\aent}}{\vek{\arel}} = 
    ( \vek{\aent}\vek{M}^T_\subj \odot \vek{\arel}\vek{M}^T_\rel) \vek{M}_\obj
\end{equation}
Notice that $\vek{x}\vek{M}^T_\subj$ are the triples with an entity in
$\vek{x}$ as their subject, $\vek{r}\vek{M}^T_\rel$ are the triples
with a relation in $\vek{r}$, and the Hadamard product is the
intersection of these.  The final multiplication by \( \vek{M}_\obj \)
finds the object entities of the triples in the intersection.  These
operations naturally extend to minibatches (see Appendix).  The reified KB has size $O(N_T)$, the sets of
triples that are intersected have size \( O(b N_T) \), and the final
result is size $O(b N_E)$, giving a final size of $O(b N_T + b N_E)$,
with no dependence on $N_R$.

Table~\ref{tab:implementations} summarizes the complexity of these
three mathematically equivalent but computationally different
implementions.  The analysis suggests that the reified KB is
preferable if there are many relations, which is the case for most
realistic KBs\footnote{The larger benchmark datasets used in this
  paper have 200 and 616 relations respectively.}.

\textbf{Distributing a large reified KB.}  The reified KB
representation is quite compact, using only six integers and three
floats for each KB triple.  However, since GPU memory is often
limited, it is important to be able to \emph{distribute} a KB across
multiple GPUs.  Although to our knowledge prior implementations of
distributed matrix operations (e.g.,
\citep{DBLP:journals/corr/abs-1811-02084}) do not support sparse
matrices, sparse-dense matrix multiplication can be distributed
 across multiple machines.  \emph{We thus implemented a
  distributed sparse-matrix implementation of reified KBs}. We
distibuted the matrices that define a reified KB ``horizontally'', so
that different triple ids $\ell$ are stored on different GPUs.
Details are provided in Appendix~\ref{sec:distrib}.

\myvspace{-0.1in}

\section{Experiments} \label{sec:expt}

\myvspace{-0.1in}

\subsection{Scalability} \label{sec:scalability}

\myvspace{-0.05in}

\begin{figure}[t]
\centering
\includegraphics[width=\textwidth]{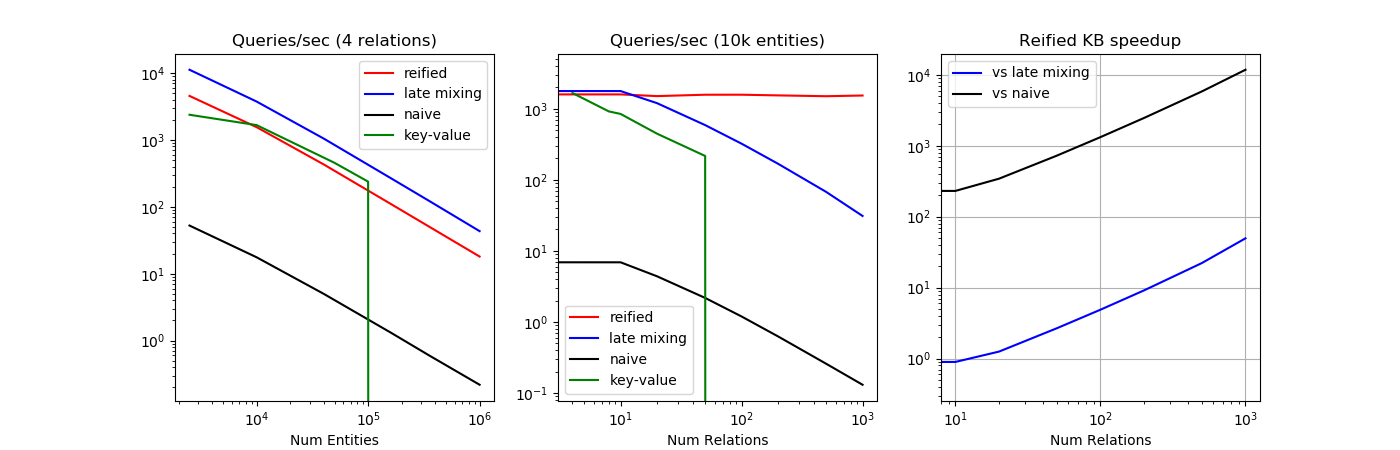}
\caption{Left and middle: inference time in queries/sec on a synthetic KB as size and number of relations is varied.  Queries/sec is given as zero
  when GPU memory of 12Gb is exceeded. Right: speedups of reified KBs over the baseline implementations.}
    \label{fig:scaling}
    \myvspace{-0.1in}
\end{figure}

Like prior work \citep{cohen2017tensorlog,de2007problog}, we used a
synthetic KB based on an $n$-by-$n$ grid to study scalability of
inference. Every grid cell is an entity, related to its immediate
neighbors via relations \textit{north}, \textit{south},
\textit{east}, and \textit{west}.  The KB for an $n$-by-$n$ grid thus
has $O(n^2)$ entities and $O(n^2)$ triples.  We measured the time
to compute the 2-hop inference \(
\follow{\follow{\vek{\aent}}{\vek{\arel}}}{\vek{\arel}} \) for
minibatches of $b=128$ one-hot vectors, and report it as queries per
second (qps) on a single GPU (e.g., qps=1280 would mean a single
minibatch requires 100ms).  We also compare to a key-value memory
network \citep{kvmem}, using an embedding size of 64 for entities and
relations, where there is one memory entry for every triple in the KB.
Further details are given in Appendix~\ref{sec:exptdetails}.

The results are shown Figure~\ref{fig:scaling} (left and middle), on a
log-log scale because some differences are very large.  With only four
relations (the leftmost plot), late mixing is about 3x faster than the
reified KB method, and about 250x faster than the naive approach.
However, for more than around 20 relations, the reified KB is faster
(middle plot).  As shown in the rightmost plot, the reified KB is 50x
faster than late mixing with 1000 relations, and nearly 12,000x faster
than the naive approach.  

With this embedding size, the speed of the key-value network is
similar to the reified KB for only four relations, however it is about
7x slower for 50 relations and 10k entities.  Additionally, the space
needed to store a triple is much larger in a key-value network than
the reified KB, so memory is exhausted when the KB exceeds 200,000
entities (with four relations), or when the KB exceeds 100 relations
(with 10,000 entities.)  The reified KB scales much better, and can
handle 10x as many entities and 20x as many relations.

\myvspace{-0.07in}

\subsection{Models using reified KBs}

\myvspace{-0.05in}

As discussed below in Section~\ref{sec:related}, the reified KB is
closely related to key-value memory networks, so it can be viewed as a
more efficient implementation of existing neural modules, optimized
for reasoning with symbolic KBs.  However, being able to include an
entire KB into a model can lead to a \emph{qualitative} difference in
model complexity, since it is not necessary to build machinery to
retrieve from the KB.  To illustrate this, below we present simple
models for several tasks, each using the reified KB in different ways,
as appropriate to the task. We consider two families of tasks:
learning semantic parsers from denotations over a large KB, and
learning to complete a KB.

\textbf{KBQA for multi-hop questions.} MetaQA
\citep{zhang2017variational} consists of 1.2M questions, evenly
distributed into one-hop, two-hop, and three-hop questions. (E.g, the
question ``\textit{who acted in a movie directed by Quentin
  Tarantino?}'' is a two-hop question.)  The accompanying KB
\citep{kvmem} contains 43k entities and 186k triples.
Past work treated one-hop, two-hop and three-hop questions separately,
and the questions are labeled with the entity ids for the ``seed
entities'' that begin the reasoning chains (e.g., the question above
would be tagged with the id of the entity for \textit{Quentin
  Tarantino}). 
  
Using a reified KB for reasoning means the neural model only needs to predict the relations used at each stage in the reasoning process.  For
each step of inference we thus compute relation sets $\vek{r}^t$ using
a differentiable function of the question, and then chain them
together with \followop{} steps. Letting $\vek{x}^0$ be the set of
entities associated with $q$, the model we use is:
$$
\textnormal{for $t=1,2,3$:}~~~
\vek{r}^t = f^{t}(q); ~~~
\vek{x}^{t} = \follow{\vek{x}^{t-1}}{\vek{r}^t}
$$ where $\follow{\vek{x}^{t-1}}{\vek{r}^t}$ is implemented with a
reified KB as described in Eq.~\ref{eq:rjoin}.  

To predict an answer on a $T$-hop subtask, we compute the softmax of
the appropriate set $\vek{x}^T$.  We used cross entropy loss of this
set against the desired answer, represented as a uniform distribution
over entities in the target set.  Each function $f^t(q)$ is a different
linear projection of a common encoding for $q$, specifically a
mean-pooling of the tokens in $q$ encoded with a pre-trained
128-dimensional word2vec model \citep{mikolov2013distributed}.  The
full KB was loaded into a single GPU in our experiments.

It is interesting to contrast this simple model with the one proposed
by \citet{zhang2017variational}.  The ``module for logic reasoning''
they propose in Section 3.4 is fairly complex, with a description that
requires a figure, three equations, and a page of text; furthermore,
training this model requires constructing an example-dependent
subgraph for each training instance.  In our model, the ``logic
reasoning'' (and all interaction with the KB) has been encapsulated
completely in the $\follow{\vek{x}}{\vek{r}}$ operation---which, as we
will demonstrate below, can be re-used for many other problems.
Encapsulating all KB reasoning with a single scalable differentiable
neural module greatly simplifies modeling: in particular, \emph{the
  problem of learning a structured KB query has been reduced to
  learning a few differentiable functions of the question}, one for
each reasoning ``hop''.  The learned functions are also interpretable:
they are mixtures of relation identifiers which correspond to soft
weighted sets of relations, which in turn softly specify which KB
relation should be used in each stage of the reasoning process.
Finally, optimization is simple, as the loss on predicted denotations
can be back-propagated to the relation-prediction functions.  

A similar modeling strategy is used in all the other models presented
below.

\textbf{KBQA on FreeBase.}  WebQuestionsSP \citep{yih2016value}
contains 4737 natural language questions, all of which are
answerable using FreeBase \citep{bollacker2008freebase}, a large
open-domain KB.  Each question $\aq$ is again labeled with the entities
$\vek{x}$ that appear in it.

FreeBase contains two kinds of nodes: real-world \textit{entities},
and \textit{compound value types} (CVTs), which represent non-binary
relationships or events (e.g., a movie release event, which includes a
movie id, a date, and a place.)  Real-world entity nodes can be
related to each other or to a CVT node, but CVT nodes are never
directly related to each other.  In this dataset, all questions can be
answered with 1- or 2-hop chains, and all 2-hop reasoning chains pass
through a CVT entity; however, unlike MetaQA, the number of hops is
not known.  Our model thus derives from $q$ three relation sets and
then uniformly mixes both potential types of inferences:

\begin{equation*}
\begin{gathered}
\vek{r}_{\textnormal{E} \rightarrow \textnormal{E}} = f_{\textnormal{E} \rightarrow \textnormal{E}}(q); ~~~
\vek{r}_{\textnormal{E} \rightarrow \textnormal{CVT}} = f_{\textnormal{E} \rightarrow \textnormal{CVT}}(q); ~~~
\vek{r}_{\textnormal{CVT} \rightarrow \textnormal{E}} = f_{\textnormal{CVT} \rightarrow \textnormal{E}}(q)
\\
\hat{\vek{a}} = \follow{\follow{\vek{x}}{\vek{r}_{\textnormal{E} \rightarrow \textnormal{CVT}}}}{\vek{r}_{\textnormal{CVT} \rightarrow \textnormal{E}}} + \follow{\vek{x}}{\vek{r}_{\textnormal{E} \rightarrow \textnormal{E}}}
\end{gathered}
\end{equation*}

We again apply a softmax to $\hat{\vek{a}}$ and use cross entropy
loss, and $f_{\textnormal{E} \rightarrow \textnormal{E}}$,
$f_{\textnormal{E} \rightarrow \textnormal{CVT}}$, and
$f_{\textnormal{CVT} \rightarrow \textnormal{E}}$ are again linear
projections of a word2vec encoding of $q$.  We used a subset of
Freebase with 43.7 million facts and 12.9 million entities, containing
all facts in Freebase within 2-hops of entities mentioned in any
question, excluding paths through some very common entities.  We split
the KB across three 12-Gb GPUs, and used a fourth GPU for the rest of
the model.

This dataset is a good illustration of the scalability issues
associated with prior approaches to including a KB in a model, such as
key-value memory networks.  A key-value network can be trained to
implement something similar to \followop{}, if it stores all the KB
triples in memory.  If we assume 64-float embeddings for the 12.9M
entities, \emph{the full KB of 43.7M facts would be 67Gb in size},
which is impractical.  Additionally performing a softmax over the
43.7M keys would be prohibitively expensive, as shown by the
experiments of Figure~\ref{fig:scaling}.  This is the reason why in
standard practice with key-value memory networks for KBs, the memory
is populated with a heuristically subset of the KB, rather than the
full KB.  We compare experimentally to this approach in
Table~\ref{tbl:exp_kbqa}.

\textbf{Knowledge base completion}.  Following
\citet{yang2017differentiable} we treat KB completion as an inference
task, analogous to KBQA: a query $q$ is a relation name and a head
entity $\vek{x}$, and from this we predict a set of tail entities.  We
assume the answers are computed with the disjunction of multiple
inference chains of varying length.  Each inference chain has a
maximum length of $T$ and we build $N$ distinct inference chains in
total, using this model (where $\vek{x}_i^0=\vek{x}$ for every chain $i$):
\begin{equation*}
\begin{gathered}
\textnormal{for $i=1,\ldots, N$ and $t = 1, \dots, T$:}~~~~
\vek{r}_i^t = f_i^t(q);~~~\vek{x}_i^{t} = \follow{\vek{x}_i^{t-1}}{\vek{r}_i^t} + \vek{x}_i^{t-1}
\end{gathered}
\end{equation*}
The final output is a
softmax of the mix of all the $\vek{x}_i^T$'s: i.e., we let \(
\hat{\vek{a}} = \textit{softmax} (\textstyle\sum_{i\in \{1\dots N\}}
\vek{x}_i^T) \).  The update \( \vek{x}_i^{t+1} =
\follow{\vek{x}_i^t}{\vek{r}_i^t} + \vek{x}_i^t \) gives the model
access to outputs of all chains of length less than $t$ (for more intuition see Appendix~\ref{sec:exptdetails}.)
The encoding
of $q$ is based on a lookup table, and each $f_i^t$ is
a learned linear transformation of $q$'s embedding.\footnote{In the
  experiments we tune the hyperparameters $T \in \{1,\ldots, 6\}$ and
  $N\in\{1,2,3\}$ on a dev set.}

\textbf{An encoder-decoder architecture for varying inferential structures.}  To explore performance on more complex reasoning tasks,
we generated simple artificial natural-language sentences describing longer chains of
relationships on a 10-by-10 grid.  For this task we used an encoder-decoder model which
emits chains of \followop{} operations.  The question is encoded with
the final hidden state of an LSTM, written here $\vek{h}^0$.  We then
generate a reasoning chain of length up to $T$ using a decoder LSTM.
At iteration
$t$, the decoder emits a scalar probability of ``stopping'', $p^t$, and
a distribution over relations to follow $\vek{r}^t$, and then, as we did for the KBQA tasks, sets \( \vek{x}^t =
\follow{\vek{x}^{t-1}}{\vek{r}^t} \). Finally the decoder updates its
hidden state to $\vek{h}^t$ using an LSTM cell that ``reads'' the ``input'' $\vek{r}^{t-1}$.  For each step $t$, the model thus contains the steps
$$
p^t = f_p(\vek{h}^{t-1});~~~
\vek{r}^t = f_r(\vek{h}^{t-1});~~~ 
\vek{x}^t = \follow{\vek{x}^{t-1}}{\vek{r}^t};~~~
\vek{h}^t = \textrm{LSTM}(\vek{h}^{t-1},\vek{r}^{t-1})
$$ 
The final
predicted location is a mixture of all the $\vek{x}_t$'s weighted by
the probability of stopping $p_t$ at iteration $t$, i.e., \(
\hat{\vek{a}} = \textit{softmax}(\sum_{t=1}^T \vek{x}^t \cdot p^t
\prod_{t'<t} (1-p^{t'})) \). The function $f_r$ is a softmax over a linear projection, and $f_p$ is a logistic function.
In the experiments, we trained on 360,000 sentences requiring between 1 and $T$ hops and tested on an additional 12,000 sentences.

\textbf{Experimental results.}  We next consider the performance of
these models relative to strong baselines for each task.  We emphasize
our goal here is \emph{not to challenge the current state of the art
on any particular benchmark}, and clearly there are many ways the
models of this paper could be improved.
(For instance, our
question encodings are based on word2vec, rather than 
contextual encodings  \citep{devlin2018bert}, and likewise relations are predicted with simple linear classifiers, rather than, say, attention queries over some semantically meaningful space,
such as might be produced with language models or KB embedding approaches 
\citep{bordes2013translating}).
Rather, our contribution is to present a generally useful scheme for
including symbolic KB reasoning into a model, and we have thus focused
on describing simple, easily understood models that do this for
several tasks.  However, it is important to confirm experimentally that the reified KB models ``work''---e.g., that they are amenable to use of standard optimizers, etc.

Performance (using Hits@1) of our models on the KBQA tasks is shown in Table
\ref{tbl:exp_kbqa}. For the non-synthetic tasks we also compare to a
Key-Value Memory Network (KV-Mem) baseline \citep{kvmem}.  For the
smaller MetaQA dataset, KV-Mem is initialized with all facts within 3
hops of the query entities, and for WebQuestionsSP it is initialized
by a random-walk process seeded by the query entities (see
\citep{graftnet,zhang2017variational} for details).  ReifKB
consistently outperforms the baseline, dramatically so for longer
reasoning chains.
The synthetic grid task shows that there is very little degradation as
chain length increases, with Hits@1 for 10 hops still 89.7\%.  It also
illustrates the ability to predict entities in a KB, as well as
relations.

We also compare these results to two much more complex architectures
that perform end-to-end question answering in the same setting used
here: VRN \citep{zhang2017variational}, GRAFT-Net \citep{graftnet}, and PullNet \citep{sun2019pullnet}.
All three systems build question-dependent subgraphs of the KB, and then
use graph CNN-like methods \citep{kipf2016semi} to ``reason'' with
these graphs.  Although not superior, ReifKB model is competitive with
these approaches, especially on the most difficult 3-hop setting.  

A small extension to this model is to mask the seed entities out of
the answers (see Appendix~\ref{sec:exptdetails}). This model (given as
ReifKB + mask) has better performance than GRAFT-Net on 2-hop
and 3-hop questions.

\begin{table}[tb]
\centering
\small
\begin{tabular}[t]{l|ccc|ccc}
\hline
          & ReifKB & ReifKB  & KV-Mem	& VRN	& GRAFT- & PullNet\\
	  & (ours) & + mask & (baseline) &      &Net & \\
\hline	           
WebQSP    & {52.7} & ---   & 46.7	& ---	& \textbf{67.8}	   &\textbf{68.1}\\		  
MetaQA    & 	   &       & 		& 	& 	 &  \\
~~1-hop   & {96.2} & ---   &  95.8	& \textbf{97.5}	& 97.0	 & 97.0  \\
~~2-hop   & 81.1 & 95.4   &  25.1	& 89.9	& 94.8 & \textbf{99.9}\\
~~3-hop   & 72.3 & 79.7  &  10.1	& 62.5	& 77.2	& \textbf{91.4}   \\
Grid      & 	   	    & 		& 	& 	 &  \\
~~5-hop   & 98.4 & ---  & ---		& ---	& ---	  & -- \\
~~10-hop  & 89.7 & ---  & ---		& ---	& ---	  & -- \\
\hline
\end{tabular}~~\begin{tabular}[t]{ll}
\hline
\multicolumn{2}{c}{non-differentiable components}\\
\multicolumn{2}{c}{of architectures}\\
\hline
KV-Mem      & initial memory\\
	    & retrieval\\
	    & \\
VRN         & question-specific \\
GRAFTNet    & subgraph retrieval \\
PullNet	    & all iterative retrievals\\
	    & \\
ReifKB(ours) & \emph{none}  \\
\hline
\end{tabular}
\caption{Hits@1 on the KBQA datasets. Results for KV-Mem and VRN on
  MetaQA are from \citep{zhang2017variational}; results for GRAFT-Net, PullNet
  and KV-Mem on WebQSP are from \citep{graftnet} and \citep{sun2019pullnet}.\label{tbl:exp_kbqa}}
\myvspace{-0.1in}
\end{table}

\begin{table}[b]
\centering
\small
\begin{tabular}[t]{r|cc}
\hline
     & \multicolumn{2}{|c}{NELL-995} \\
     & H@1 & H@10 \\
     \hline
ReifKB (Ours)& 64.1 & 82.4\\
\hline
DistMult*& 61.0 & 79.5 \\
ComplEx* & 61.2 & 82.7 \\
ConvE*   & \textbf{67.2} & \textbf{86.4} \\
\hline
\end{tabular}~~~\begin{tabular}[t]{l|cc}
\hline
     & ReifKB (Ours) & MINERVA \\
     \hline
NELL-995              & 64.1 & \textbf{66.3}  \\
Grid with seed entity & & \\
~~10-hop NSEW      & 98.9 & \textbf{99.3}  \\
~~10-hop NSEW-VH   & \textbf{73.6} & 34.4  \\
MetaQA 3-hop          & \textbf{72.3} & 41.7  \\
\hline
\end{tabular}
\caption{{Left:} Hits@1 and Hits@10 for KB completion on NELL
  995.  Starred KB completion methods are transductive, and do not
  generalize to entities not seen in training. {Right:}
  Comparison to MINERVA on several tasks for Hits@1.}
\label{tbl:exp_kbc}
\myvspace{-0.1in}
\end{table}

For KB completion, we evaluated the model on the NELL-995 dataset
\citep{xiong2017deeppath} which is paired with a KB with 154k facts,
75k entities, and 200 relations.  On the left of
Table~\ref{tbl:exp_kbc} we compare our model with three popular
embedding approaches (results are from \citet{das2017go}).  The
reified KB model outperforms DistMult \citep{yang2014embedding}, is
slightly worse than ConvE \citep{dettmers2018convolutional}, and is
comparable to ComplEx \citep{trouillon2017knowledge}.  

The competitive performance of the ReifKB model is perhaps surprising,
since it has many fewer parameters than the baseline models---only one
float and two integers per KB triple, plus a small number of
parameters to define the $f_i^t$ functions for each relation.  The
ability to use fewer parameters is directly related to the fact that
our model \emph{directly uses inference on the existing symbolic KB}
in its model, rather than having to learn embeddings that approximate
this inference.  Or course, since the KB is incomplete, some learning
is still required, but learning is quite different: the system learns
logical inference chains in the incomplete KB that approximate a
target relation.  In this setting for KBC, the ability to perform
logical inference ``out of the box'' appears to be very advantageous.

Another relative disadvantage of KB embedding methods is that KB
embeddings are generally \trm{transductive}---they only make
predictions for entities seen in training.  As a non-transductive
baseline, we also compared to the MINERVA model, which uses
reinforcement learning (RL) methods to learn how to traverse a KB to
find a desired answer.  Although RL methods are less suitable as
``neural modules'', MINERVA is arguably a plausible competitor to
end-to-end learning with a reified KB.  

MINERVA slightly outperforms our simple KB completion model on the
NELL-995 task. However, unlike our model, MINERVA is trained to find a
\emph{single answer}, rather than trained to infer a \emph{set of
  answers}.  To explore this difference, we compared to MINERVA on the
grid task under two conditions: (1) the KB relations are the grid
directions north, south, east and west, so the output of the target chain is always a \emph{single} grid location, and (2) the KB
relations also include a ``vertical move'' (north or south) and a
``horizontal move'' (east or west), so the result of the target chain
can be a \emph{set} of locations.  As expected MINERVA's performance
drops dramatically in the second case, from 99.3\% Hits@1 to 34.4 \%,
while our model's performance is more robust.  MetaQA answers can also
be sets, so we also modified MetaQA so that MINERVA could be used (by
making the non-entity part of the sentence the ``relation'' input and
the seed entity the ``start node'' input) and noted a similarly poor
performance for MINERVA. These results are shown on the right of
Table~\ref{tbl:exp_kbc}.


In Tables~\ref{tbl:exe_time} we compare the training time of our model
with minibatch size of 10 on NELL-995, MetaQA, and
WebQuestionsSP. With over 40 million facts and nearly 13 million
entities from Freebase, it takes less than 10 minutes to run one epoch
over WebQuestionsSP (with 3097 training examples) on four P100 GPUs.
In the accompanying plot, we also summarize the tradeoffs between
accuracy and training time for our model and three baselines on the
MetaQA 3-hop task. (Here ideal performance is toward the upper left of
the plot).  The state-of-the-art PullNet \cite{sun2019pullnet} system, which uses a learned
method to incrementally retrieve from the KB, is about 15 times slower
than the reified KB system.  GRAFT-Net is only slightly less accurate, but
also only slightly faster: recall that GRAFT-Net uses a
heuristically selected subset (of up to 500 triples) from the KB for
each query, while our system uses the full KB.  Here the full KB is about 400
times as large as the question-specific subset used by GRAFT-Net.  
A key-value memory baseline including the full KB is
nearly three times as slow as our system, while also performing quite
poorly.

\begin{table}[t]
\centering
\small
\begin{tabular}[b]{lccc}
\hline
               & NELL-995 & MetaQA-3hop  & WebQuestionsSP \\
               \hline
\# Facts        & 154,213 & 196,453 & 43,724,175     \\
\# Entities     & 75,492  & 43,230  & 12,942,798     \\
\# Relations    & 200     & 9       & 616            \\ \hline \hline
Time (seconds) & 44.3    & 72.6    & 1820          \\ \hline
 & & & 
\end{tabular}~\includegraphics[width=0.3\textwidth]{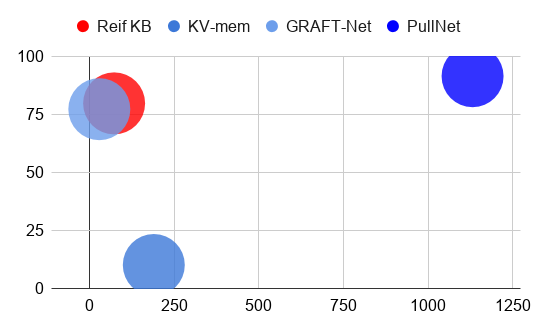}
\caption{Left, time to run 10K examples for KBs of different size.
  Right, time for 10k examples vs~ Hits@1 performance for ReifKB
  compared to three baselines on MetaQA-3hop questions.}
 \label{tbl:exe_time}
\myvspace{-0.25in}
\end{table}
\myvspace{-0.1in}

\section{Related Work} \label{sec:related}

\myvspace{-0.1in}

The \followop{} operation using reified KBs is implemented in an
open-source package called NQL, for neural query language. NQL
implements a broader range of operations for manipulating KBs, which
are described in a companion paper \citep{nqlarxiv}.  This
paper focuses on implementation and evaluation of the \followop{}
operation with different KB representations, issues not covered in the companion paper.

TensorLog \citep{cohen2017tensorlog}, a probabilistic logic which also
can be compiled to Tensorflow, and hence is another differentiable
approach to neuralizing a KB.  TensorLog is also based on sparse
matrices, but does not support relation sets, making it unnatural to
express the models shown in this paper, and does not use the more
efficient reified KB representation.  The differentiable theorem prover (DTP) is another differentiable
logic \citep{rocktaschel2017end}, but DPT appears to be much less
scalable: it has not been applied to KBs larger than a few thousand
triples.  The Neural ILP system \citep{yang2017differentiable} uses
approaches related to late mixing together with an LSTM controller to
perform KB completion and some simple QA tasks, but it is a monolithic
architecture focused on rule-learning, while in contrast we propose a
re-usable neural component, which can be used in as a component in
many different architectures, and a scalable implementation of this.
It has also been reported that neural ILP does not scale to the size of the
NELL995 task \citep{das2017go}.

The goals of this paper are related to KB embedding methods, but
distinct. In KB embedding, models are generally fully differentiable,
but it is not considered necessary (or even desirable) to accurately
match the behavior of inference in the original KB.  Being able to
construct a learned \emph{approximation} of a symbolic KB is
undeniably useful in some contexts, but embedded KBs also have many
disadvantages.  In particular, they are much larger than a reified KB, with many more
learned parameters---typically a long dense vector for every KB
entity.  
Embedded models are typically evaluated by their ability to score a
single triple accurately, and many models are not capable of executing
multi-step KB inferences efficiently; further, models that do allow
multi-step inference are known to produce cascaded errors on long
reasoning chains \citep{guu2015traversing,Hamilton2018embedding}.  In
contrast we focus on accurate models of reasoning in a symbolic KB,
which requires consideration of novel scalability issues associated
with sparse matrice representations.

Mathematically, our definition of relation-set following is much like
the bilinear model for path following from \citet{guu2015traversing};
however, we generalize this to path queries that include weighted sets
of relations, allowing the relations in paths to be learned.  Similar
differences apply to the work of \citet{Hamilton2018embedding}, which
extends the work of \citet{guu2015traversing} to include intersection
operations.  The vector representation used here for weighted sets in a
reified KB makes intersection trivial to implement, as intersection
corresponds to Hadamard product.  Conveniently set union also
corresponds to vector sum, and the complement of $\sent$ is
$1-\vek{\aent}$, which is perhaps why only a single additional neural
operation is needed to support the KB reasoning tasks needed for the
five benchmark tasks considered here.

Neural architectures like memory networks \citep{weston2014memory}, or
other architectures that use attention over some data structure
approximating assertions
\citep{andreas2016neural,DBLP:journals/corr/abs-1808-09942} can be
used to build soft versions of \followop{}: however, they also do not
scale well to large KBs, so they are typically used either with a
non-differentiable \textit{ad hoc} retrieval mechanism, or else in
cases where a small amount of information is relevant to a question
\citep{weston2015towards,zhong2017seq2sql}.  Similarly graph
CNNs \citep{kipf2016semi} also can be used for reasoning, and often do
use sparse matrix multiplication, but again existing implementations
have not been scaled to tens of millions of triples/edges or millions
of entities/graph nodes.  Additionally, while graph CNNs have been used
for reasoning tasks, the formal connection between them and logical
reasoning remains unclear, whereas there is a precise connection
between \followop{} and inference.

Reinforcement learning (RL) methods have been used to learn mappings from
natural-language questions to non-differentiable logical
representations \citep{liang2016neural,liang2018memory} and have also
been applied to KB completion tasks
\citep{das2017go,xiong2017deeppath}.  Above we compared experimentally
to MINERVA, one such method; however, the gradient-based approaches
enabled by our methods are generally preferred as being easier to
implement and tune on new problems, and easier to combine in a modular
way with other architectural elements.

\myvspace{-0.1in}

\section{Conclusions}

\myvspace{-0.1in}

We introduced here a novel way of representing a symbolic knowledge
base (KB) called a sparse-matrix reified KB.  This representation
enables neural modules that are fully differentiable, faithful to the
original semantics of the KB, expressive enough to model multi-hop
inferences, and scalable enough to use with realistically large KBs.
In a reified KB, all KB relations are represented with three sparse
matrices, which can be distributed across multiple GPUs, and symbolic
reasoning on realistic KBs with many relations is much faster than
with naive implementations---more than four orders of magnitude faster
on synthetic-data experiments compared to naive sparse-matrix
implementations.

This new architectural component leads to radically simpler
architectures for neural semantic parsing from denotations and KB
completion---in particular, they make it possible to learn neural KBQA
models in a completely end-to-end way, mapping from text to KB entity
sets, for KBs with tens of millions of triples and entities and
hundreds of relations.


\subsubsection*{Acknowledgments}  The authors are greatful to comments and suggestions from Fernando Peireira, Bhuwan Dhingra, and many other colleagues on earlier versions of this work.

\newpage

\bibliographystyle{iclr2020_conference}
\bibliography{./iclr}

\newpage

\appendix

\section{Additional background and extensions} \label{sec:types}

\textbf{KBs, entities, and relations, and types.}  In the more general
case, a KB consists of \trm{entities}, \trm{relations}, and
\trm{types}.  Again use $\aent$ to denote an entity and $\arel$ to
denote a relation.  We also assume each entity $\aent$ has a
\trm{type}, written $\ty(\aent)$, and let $N_\aty$ denote the number
of entities of type $\aty$.  Each entity $\aent$ in type $\aty$ has a
unique index $\aind_\aty(\aent)$, which is an integer between $1$ and
$N_{\aty}$.  We write $\aent_{\aty,i}$ for the entity that has index
$i$ in type $\aty$, or $\aent_i$ if the type is clear from context.

Every relation $\arel$ has a \trm{subject type} $\aty_\subj$ and an
\trm{object type} $\aty_\obj$, which constrain the types of $\aent$
and $\aent'$ for any pair $(\aent,\aent') \in \arel$.  Hence $\arel$
can be encoded as a subset of \( \{1,\ldots,N_{\aty_\subj}\} \times
\{1,\ldots,N_{\aty_\obj}\} \).  Relations with the same subject and
object types are called \textit{type-compatible}.

Our differentiable operations are based on \trm{typed weighted sets},
where again each element $\aent$ of weighted set $\sent$ is associated
with a non-negative real number, written $\setw{\aent}{\sent}$, and we
define $\setw{\aent}{\sent}\equiv{}0$ for all $\aent\not\in\sent$.  A set $\sent$ has a type $\ty(\sent)=\aty$, and all members of $\sent$must be entities
of type $\aty$.

We also assume every relation $\arel$ in a KB is associated with an
entity $\aent_\arel$, and hence, an index and a type. Sets of
relations $\srel$ are allowed only if all members are type-compatible.  For
example $\srel = \{ \textit{writer\_of}, \textit{director\_of} \}$
might be a set of type-compatible relations.

A weighted set $\sent$ of type $\aty$ can be encoded as an
\trm{entity-set vector} $\vek{\aent} \in \aR^{N_\aty}$, where the
$i$-th component of $\vek{\aent}$ is the weight of the $i$-th entity
of that type in the set $\sent$: e.g., \(
\vek{\aent}[\aind_\aty(\aent)] = \setw{\aent}{\sent} \). We also use
$\ty(\vek{\aent})$ to denote the type $\aty$ of the set encoded by
$\vek{\aent}$.

A relation $\arel$
with subject type $\aty_1$ and object type $\aty_2$ can be encoded as
a \trm{relation matrix} $\vek{M}_\arel \in \aR^{N_{\aty_1} \times
  N_{\aty_2}}$.

\textbf{Background on sparse matrices.} A COO encoding consists of a $N_\arel \times 2$ matrix
$\vek{Ind}_\arel$ containing pairs of entity indices, and a parallel
vector $\vek{w}_\arel \in \aR^{N_\arel}$ containing the weights of the
corresponding entity pairs.  In this encoding, if $(i,j)$ is row $k$
of $\vek{Ind}$, then $\vek{M}_\arel[i,j] = \vek{w}_\arel[k]$, and if
$(i,j)$ does not appear in $\vek{Ind}_\arel$, then $\vek{M}[i,j]$ is
zero.

\textbf{Extension to soft KBs}. In the paper, we assume the non-zero weights in a relation matrix $\vek{M}_r$ are all equal to 1.0.  This can be relaxed: if assertions in a KB are associated with confidences, then 
this confidence can be stored in $\vek{M}_\arel$.  
In this case, the reified KB must be extended to encode the weight for a triple: 
we find it convenient to redefine $\vek{M}_\rel$ to hold that weight.  In particular if the weight for the
the $\ell$-th triple $r_k(x_i,x_j)$ is $w_\ell$, then we let
$$
\vek{M}_\rel[\ell,m] \equiv \left\{
   \begin{array}{l}
     \mbox{$w_\ell$~~if $m=k_\ell$} \\
     \mbox{$0$~~else} \\
   \end{array}
\right.
$$

\section{Proof of Claim 1} \label{sec:proof}

\setcounter{claim}{0}

\begin{claim} The support of $\follow{\vek{\aent}}{\vek{\arel}}$ is
exactly the set of $\srel$-neighbors($\sent$).
\end{claim}

To better understand this claim, let $\vek{z}=\follow{\vek{\aent}}{\vek{\arel}}$.  The claim
states $\vek{z}$ can approximate the $\srel$ neighborhood of any
hard sets $\srel,\sent$ by setting to zero the appropriate components
of $\vek{\aent}$ and $\vek{\arel}$.  It is also clear that
$\vek{z}[j]$ decreases when one decreases the weights in
$\vek{\arel}$ of the relations that link $\aent_j$ to entities in
$\sent$, and likewise, $\vek{z}[j]$ decreases if one
decreases the weights of the entities in $\sent$ that are linked to
$\aent_j$ via relations in $\srel$, so there is a
smooth, differentiable path to reach this approximation.

More formally, consider first a matrix $\vek{M}_\arel$ encoding a
single binary relation $\arel$, and consider the vector
$\vek{\aent}' = \vek{\aent} \vek{M}_\arel$.  As weighted sets,
$\sent$ and $r$ have non-negative entries, so clearly for all $i$,
$$ \vek{\aent}'[j] \not= 0 \mbox{~~iff~~} 
 \exists j : \vek{M}_\arel[i,j] \not= 0 \wedge \vek{\aent}[i] \not= 0 
 \mbox{~~iff~~} \exists \aent_i\in \sent \mbox{~so that~} (\aent_i,\aent_j) \in \arel
$$ and so if $\vek{\arel}$ is a one-hot vector for the set
 $\{\arel\}$, then the support of $\follow{\vek{\aent}}{\vek{\arel}}$ is
 exactly the set $\arel$-neighbors($\sent$).  Finally note that the
 mixture $\vek{M}_\srel$ has the property that
 $\vek{M}_\srel[i(e_1),i(e_2)] > 0$ exactly when $e_1$ is related to
 $e_2$ by some relation $\arel \in \srel$.

\section{Minibatched computations of naive and late mixing}
\label{sec:mbatch}

The major problem with naive mixing is that, in the absence of general
sparse tensor contractions, it is difficult to adapt to
mini-batches---i.e., a setting in which $\vek{x}$ and $\vek{r}$ are
replaced with matrices $\vek{X}$ and $\vek{R}$ with minibatch size
$b$.  An alternative strategy is \trm{late mixing}, which mixes the
\emph{output} of many single-relation following steps, rather than
mixing the KB itself:
$$
    \follow{\vek{X}}{\vek{R}} = 
       \sum_{k=1}^{N_R} (\vek{R}[:,k] \cdot \vek{X} \vek{M}_k)
$$
Here \(\vek{R}[:,k]\), the $k$-th column of $\vek{R}$, is
``broadcast'' to element of the matrix $\vek{X} \vek{M}_k$.  As noted
in the body of the text, while there are $N_R$ matrices $\vek{X}
\vek{M}_k$, each of size $O(b N_E$), they need not all be stored at
once, so the space complexity becomes $O(b N_E + b N_R + N_T)$;
however we must now sum up $N_R$ dense matrices.

The implementation of \followop{} for the reified KB can be
straightforwardedly extended to a minibatch:
$$
    \follow{\vek{X}}{\vek{R}} = 
    ( \vek{X}\vek{M}^T_\subj \odot \vek{R}\vek{M}^T_\rel) \vek{M}_\obj
$$

\section{Distributed matrix multiplication} \label{sec:distrib}

Matrix multiplication $\vek{x}\vek{M}$ was distributed as follows:
$\vek{x}$ can be split into a ``horizontal stacking'' of $m$
submatrices, which we write as \( \left[ \vek{x}_1 ; \ldots ;\vek{x}_m
  \right] \), and $\vek{M}$ can be similarly partitioned into $m^2$
submatrices.  We then have the result that
$$ \vek{x} \vek{M} = 
\left[ \vek{x}_1 ; \vek{x}_2; \ldots  ;\vek{x}_m \right] 
\left[ 
    \begin{array}{cccc}
    \vek{M}_{1,1} & \vek{M}_{1,2} & \ldots & \vek{M}_{1,m}\\
    \vdots & \vdots &  & \vdots \\
    \vek{M}_{m,1} & \vek{M}_{m,2} & \ldots & \vek{M}_{m,m}\\
    \end{array} \right] =
    \left[
      (\sum_{i=1}^m \vek{x}_1 \vek{M}_{i,1}); \ldots; 
      (\sum_{i=1}^m \vek{x}_m \vek{M}_{i,m})
    \right]
$$ 
This can be computed without storing either $\vek{X}$ or $\vek{M}$ on
a single machine, and mathematically applies to both dense and sparse
matrices. In our experiments we distibuted the matrices that define a
reified KB ``horizontally'', so that different triple ids $\ell$ are
stored on different GPUs.

Specifically, we shard the ``triple index'' dimension $N_T$ of matrices $\vek{M}_\subj$, $\vek{M}_\rel$ and $\vek{M}_\obj$ in Eq. \ref{eq:rjoin} to perform a distributed \followop{} on the reified KB. Let $\vek{M}_{\subj, i}$ be the $i$'th shard of matrix $\vek{M}_\subj$, and thus $\vek{M}_\subj = [\vek{M}_{\subj, 1}^T; \ldots; \vek{M}_{\subj, m}^T]^T \in \aR^{N_T \times N_E}$. $\vek{M}_\obj$ and $\vek{M}_\rel$ are represented in the similar way. A distributed \followop{} is computed as a combination of \followop{} results on all shards of the KB.
\begin{align}
    \follow{\vek{\aent}}{\vek{\arel}} 
    &= ( \vek{\aent}\vek{M}^T_\subj \odot \vek{\arel}\vek{M}^T_\rel) \vek{M}_\obj \nonumber \\
    &= \left([\vek{\aent} \vek{M}^T_{\subj, 1}; \ldots; \vek{\aent} \vek{M}^T_{\subj, m}] \odot [\vek{\arel} \vek{M}^T_{\rel, 1}; \ldots; \vek{\arel} \vek{M}^T_{\rel, m}] \right) 
    \left[ 
    \begin{array}{c}
    \vek{M}_{\obj,1}\\
    \vdots\\
    \vek{M}_{\obj,m}\\
    \end{array} \right]\\
    &= \sum_{i=1}^m ( \vek{\aent}\vek{M}^T_{\subj, i} \odot \vek{\arel}\vek{M}^T_{\rel, i}) \vek{M}_{\obj, i}
\end{align}

This method can be easily extended to a mini-batch of examples $\vek{X}$.

\section{Experimental Details} \label{sec:exptdetails}

\textbf{Reproducing experiments.}
To reproduce these experiments, first download and install the Google \texttt{language} package\footnote{\texttt{https://github.com/google-research/language.git}}.
Many of the experiments in this paper can be reproduced using scripts stored in the some subdirectory of the 
source directory \texttt{language/nql/demos}: for example, the scalability experiments of Figure~\ref{fig:scaling} can be performed using scripts in
\texttt{language/nql/demos/gridworld\_scaling/}.

\textbf{Grid experiments.}  In the grid experiments, the entity vector
$\vek{\aent}$ is a randomly-chosen singleton set, and the relation vector
$\vek{\arel}$ weights relations roughly uniformly---more specifically, each relation has weight 1+$\epsilon$ where $\epsilon$ is a drawn uniformly at random between $0$ and $0.001$.\footnote{If the relation weights do not vary from trial to trial, some versions of Tensorflow will optimize computation by precomputing and caching the matrix $\vek{M}_R$ from Eq.~\ref{eq:MR}, which speeds up the naive method considerably.  Of course, this optimization is impossible when learning relation sets.} 
We vary the number of relations by inventing $m$ new relation names and assigning existing grid edges to each new relation.  These experiments were conducted on a Titan Xp GPU with 12Gb of memory.

For key-value networks, the key is the concatenation of a relation and
a subject entity, and the value is the object entity.  We considered
only the run-time for queries on an untrained randomly-initialized
network (since run-time performance on a trained network would be the
same); however, it should be noted that considerable time that might
be needed to train the key-value memory to approximate the KB. (In fact, it is not obvious under what conditions a KB can be approximated well by the key-value memory.)


We do not show results on the grid task for smaller minibatch sizes,
but both reified and late mixing are about 40x slower with $b=1$ than
with $b=128$.

\textbf{WebQuestionsSP experiments.}  For efficiency, on this problem
we exploit the type structure of the problem (see
Appendix~\ref{sec:types}). Our model uses two types of nodes, CVT and
entity nodes.  The model also uses three types of relations: relations
mapping entities to entities, relations mapping entities to CVT nodes;
and relations mapping CVT nodes to entity nodes.

\textbf{MetaQA experiments.}  An example of a 2-hop question
in MetaQA could be ``Who co-starred with Robert Downey Jr. in their
movies?'', and the answer would be a set of actor entities, e.g., ``Chris
Hemsworth'', ``Thomas Stanley'', etc. Triples in the knowledge base are
represented as (subject, relation, object) triples, e.g., (``Robert Downey
Jr.'', ``act\_in'', ``Avengers: Endgame''), (``Avengers: Endgame'', ``stars'', ``Thomas
Stanley''), etc.  The quoted strings here all indicate KB entities.

We also observed that in the MetaQA
2-hop and 3-hop questions, the questions often exclude the seed
entities (e.g., ``other movies with the same director as Pulp
Fiction''). This can be modeled by masking out seed entities from the
predictions after the second hop (ReifKB + mask in the table).

\textbf{Timing on MetaQA and other natural problems.}  The raw data
for the bubble plot of Table~\ref{tbl:exe_time} is below.

\medskip

\centerline{
\begin{tabular}{rcc}
\hline
Time (seconds)	& Accuracy (hits@1)	& Method \\ \hline
72.6	& 79.7	& Reif KB \\ 
189.8	& 10.1	& KV-mem \\
28.9	& 77.2	& GRAFT-Net \\
1131.0	& 91.4	& PullNet\\
\hline
\end{tabular}}

\textbf{Discussion of the KB completion model.}  The KB completion
model is
$$
\begin{gathered}
\textnormal{for $i=1,\ldots, N$ and $t = 1, \dots, T$:}~~~~
\vek{r}_i^t = f_i^t(q);~~~\vek{x}_i^{t} = \follow{\vek{x}_i^{t-1}}{\vek{r}_i^t} + \vek{x}_i^{t-1}
\end{gathered}
$$
It may not be immediately obvious why we used 
$$
\vek{x}_i^{t} = \follow{\vek{x}_i^{t-1}}{\vek{r}_i^t} + \vek{x}_i^{t-1} 
$$
instead of the simpler 
$$
\vek{x}_i^{t} = \follow{\vek{x}_i^{t-1}}{\vek{r}_i^t}
$$ In the main text, we say that this ``gives the model access to
outputs of all chains of length less than $t$''.  This statement is probably
easiest to understand by considering a concete example.
Let us simplify notation slightly by
dropping the subscripts and writing
$\follow{\vek{x}_i^{t-1}}{\vek{r}_i^t}$ as $f^t(\vek{x}^{t-1})$.  Now
expand the definition of $\vek{x}^{t}$ for a few small values of $t$,
using the linearity of the definition of \followop{} where appropriate to
simplify:
\begin{eqnarray*}
\vek{x}^1 & = & f^1(\vek{x}^0) + \vek{x}^0 \\
\vek{x}^2 & = & f^2(\vek{x}^1) + \vek{x}^1 \\
          & = & f^2\big((f^1(\vek{x}^0) + \vek{x}^0\big) + \big((f^1(\vek{x}^0) + \vek{x}^0\big) \\
          & = & f^2(f^1(\vek{x}^0)) + f^2(\vek{x}^0) + f^1(\vek{x}^0) + \vek{x}^0 \\
\vek{x}^3 & = & f^3(\vek{x}^2) + \vek{x}^2 \\
          & = & f^3\big((f^2(f^1(\vek{x}^0)) + f^2(\vek{x}^0) + f^1(\vek{x}^0) + \vek{x}^0\big) + f^2(f^1(\vek{x}^0)) + f^2(\vek{x}^0) + f^1(\vek{x}^0) + \vek{x}^0 \\
          & = & f^3(f^2(f^1(\vek{x}^0))) + f^3(f^2(\vek{x}^0)) + f^3(f^1(\vek{x}^0)) + f^3(\vek{x}^0) + f^2(f^1(\vek{x}^0)) + f^2(\vek{x}^0) + f^1(\vek{x}^0) + \vek{x}^0 \\
\end{eqnarray*}
A pattern is now clear: with this recursive definition $\vek{x}^t$
expands to a mixture of many paths, each of which applies a different
subset of $f^1$, \ldots, $f^t$ to the initial input $\vek{x}$. Since
the weights of the mixture can to a large extent be controlled by
varying the norm of the relation vectors $\vek{r}^1$,
\ldots,$\vek{r}^t$, this ``kernel-like trick'' increases the
expressive power of the model without introducing new parameters.  The
final mixture of the $\vek{x}^t$'s seems to provide a bias towards
accepting the output of shorter paths, which appears to be useful in
practice.
\end{document}